\documentclass[acmsmall]{acmart}

\usepackage{booktabs} 
\usepackage[linesnumbered,boxed,ruled,vlined]{algorithm2e}
\usepackage{amsmath,bm}
\usepackage{array}
\usepackage{graphicx, subfigure}
\usepackage{url,soul}
\usepackage{enumitem}
\usepackage{multirow,array}
\usepackage{color}
\usepackage{mathrsfs}
\usepackage{hyperref}
\usepackage[perpage,symbol]{footmisc}

\newcommand{\modelname}{SPNN}
\newcommand{\nosection}[1]{\vspace{2pt}\noindent\textbf{#1.}}

\usepackage{dsfont,amssymb}
\usepackage{listings}

\definecolor{lightgray}{rgb}{0.95, 0.95, 0.95}
\definecolor{darkgray}{rgb}{0.4, 0.4, 0.4}
\definecolor{purple}{rgb}{0.65, 0.12, 0.82}
\definecolor{ocherCode}{rgb}{1, 0.5, 0} 
\definecolor{blueCode}{rgb}{0, 0, 0.93} 
\definecolor{greenCode}{rgb}{0, 0.6, 0} 

\lstdefinelanguage{cement}{
    sensitive=true,
    keywords={%
    import, for, in, sess, def, return, super, __init__
    },
    ndkeywords={matmul, softmax, init, placeholder, random_uniform, softmax_cross_entropy_with_logits, reduce_mean, forward, backward, self,  execute_algorithm_2, update_client_models}, 
    comment=[l]{\#},
    morestring=[b]',
    morestring=[b]"
}
\setcopyright{acmcopyright}
\acmJournal{TIST}
\acmYear{2021} \acmVolume{1} \acmNumber{1} \acmArticle{1} \acmMonth{1} \acmPrice{15.00}\acmDOI{10.1145/3501809}

\begin{document}
\title[Privacy-Preserving Deep Neural Network]{Towards Scalable and Privacy-Preserving Deep Neural Network via Algorithmic-Cryptographic Co-design
}
\author{Jun Zhou}
\affiliation{%
  \institution{College of Computer Science, Zhejiang University}
  \city{Hangzhou}
  \country{China}
}
\affiliation{%
  \institution{Ant Group}
  \city{Hangzhou}
  \country{China}
}

\author{Longfei Zheng}
\affiliation{%
  \institution{Ant Group}
  \city{Hangzhou}
  \country{China}
}

\author{Chaochao Chen*}\thanks{*Corresponding author. }
\affiliation{%
  \institution{College of Computer Science, Zhejiang University}
  \city{Hangzhou}
  \country{China}
}
\email{zjuccc@zju.edu.cn}

\author{Yan Wang}
\affiliation{%
  \institution{School of Computing, Macquarie University}
  \city{Sydney}
  \country{Australia}
}

 \author{Xiaolin Zheng}
 \affiliation{%
   \institution{College of Computer Science, Zhejiang University}
   \city{Hangzhou}
   \country{China}
 }
\affiliation{%
\institution{JZTData Technology}
\city{Hangzhou}
\country{China}
}

\author{Bingzhe Wu}
\affiliation{%
  \institution{Peking University}
  \city{Beijing}
  \country{China}
}

\author{Cen Chen}
\affiliation{%
  \institution{East China Normal University}
  \city{Shanghai}
  \country{China}
}

\author{Li Wang}
\affiliation{%
  \institution{Ant Group}
  \city{Hangzhou}
  \country{China}
}

 \author{Jianwei Yin}
 \affiliation{%
   \institution{College of Computer Science, Zhejiang University}
   \city{Hangzhou}
   \country{China}
 }
\renewcommand{\shortauthors}{Jun Zhou, Longfei Zheng, and Chaochao Chen et al.}

\begin{abstract}
Deep Neural Networks (DNNs) have achieved remarkable progress in various real-world applications, especially when abundant training data are provided. 
However, data isolation has become a serious problem currently. 
Existing works build privacy preserving DNN models from either \textit{algorithmic} perspective or \textit{cryptographic} perspective. 
The former mainly splits the DNN computation graph between data holders or between data holders and server, which demonstrates good scalability but suffers from accuracy loss and potential privacy risks.
In contrast, the latter leverages time-consuming cryptographic techniques, which has strong privacy guarantee but poor scalability.
In this paper, we propose \modelname ~ --- ~ a \underline{S}calable and \underline{P}rivacy-preserving deep \underline{N}eural \underline{N}etwork learning framework, from algorithmic-cryptographic co-perspective. 
From algorithmic perspective, we split the computation graph of DNN models into two parts, i.e., the private data related computations that are performed by data holders and the rest heavy computations that are delegated to a semi-honest server with high computation ability. 
From cryptographic perspective, we propose using two types of cryptographic techniques, i.e., secret sharing and homomorphic encryption, for the isolated data holders to conduct private data related computations privately and cooperatively. 
Furthermore,
we implement \modelname~in a decentralized setting and introduce user-friendly APIs. 
Experimental results conducted on real-world datasets demonstrate the superiority of our proposed \modelname. 
\end{abstract}

%
%

\ccsdesc[500]{Security and privacy~Privacy protections}
\ccsdesc[500]{Security and privacy~Usability in security and privacy}
\ccsdesc[500]{Computing methodologies~Machine learning}

%
%

\keywords{Privacy preserving, secret sharing, homomorphic encryption, deep neural network}


\maketitle


\section{Introduction}\label{sec-intro}
\begin{figure*}
\centering
\subfigure [\emph{Plaintext DNN}]{ \label{compare-exam-nn} \includegraphics[height=3.5cm]{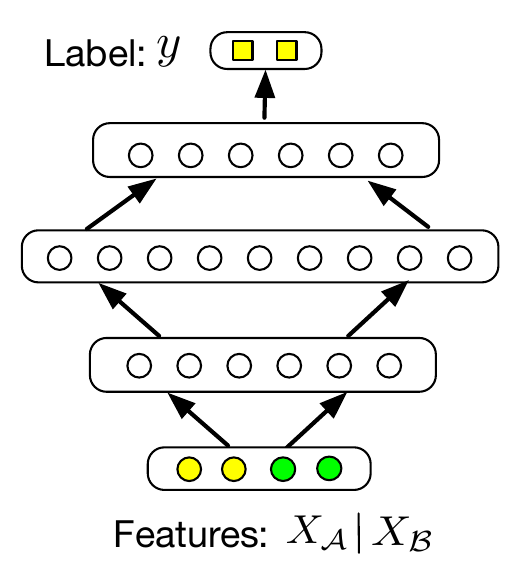}} \hspace{.5cm}
\subfigure[\emph{Algorithmic DNN \cite{vepakomma2018split}}] { \label{compare-exam-split} \includegraphics[height=3.5cm]{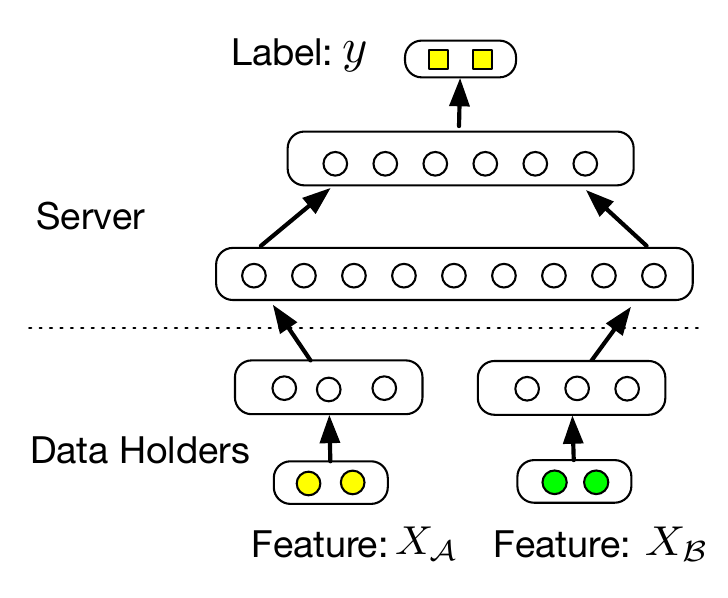}} \hspace{.5cm}
\subfigure [\emph{Cryptographic DNN \cite{mohassel2017secureml}}]{ \label{compare-exam-sml} \includegraphics[height=3.5cm]{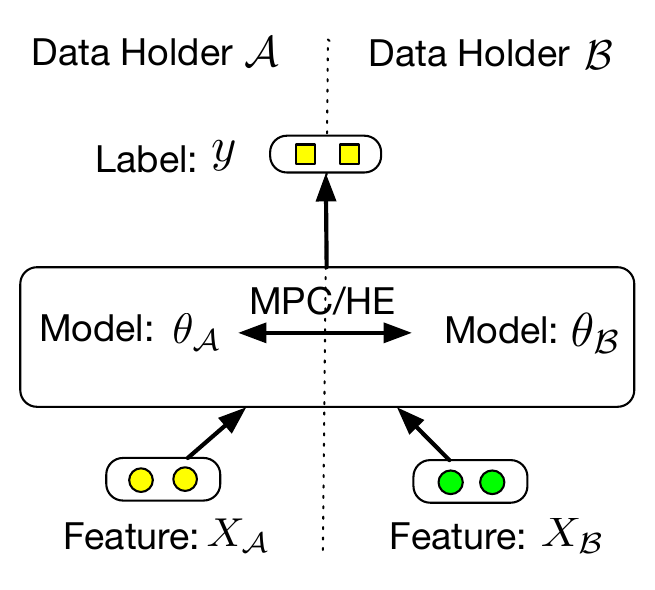}}
\caption{Comparison of existing approaches. Here, we assume there are only two data holders ($\mathcal{A}$ and $\mathcal{B}$), $\mathcal{A}$ has partial feature ($X_\mathcal{A}$, shown in yellow dots) and label ($y$, shown in yellow squares), and $\mathcal{B}$ has partial feature ($X_\mathcal{B}$, shown in green dots). }
\label{compare-exam}
\end{figure*}

Deep Neural Networks (DNN) have achieved remarkable progresses in various applications such as computer vision \cite{howard2017mobilenets}, sales forecasting \cite{chen2019much}, fraud detection \cite{Liu2020alle,dou2020enhan}, and recommender system \cite{zhu2019dtcdr,chen20tist}, due to its powerful ability to learn hierarchy  representations \cite{lecun2015deep}.
Training a good DNN model often requires a large amount of data. However, in practice, integrated data are always held by different parties. 
Traditionally, when multiple parties want to build a DNN model (e.g., a fraud detection model) together, they need to aggregate their data and train the model with the plaintext data, as is shown in Figure \ref{compare-exam-nn}. 
With all kinds of national data protection regulations coming into force, data isolation has become a serious problem currently. 
As a result, different organizations (data holders) are reluctant to or cannot share sensitive data with others due to such regulations and they have to train DNN models using their own data. 
To this end, each data holder has to train DNN models using its own data. 
Therefore, 
such a \textit{data isolation problem} has constrained the power of DNN, since DNN usually achieves better performance with more high-quality data. 

\subsection{Existing methods and Shortcomings}
Existing researches solve the above problem from either \textit{algorithmic} perspective or \textit{cryptographic} perspective. 

\textbf{Algorithmic methods} build privacy preserving DNN by spliting the computation graphs of DNN from an algorithmic perspective \cite{gupta2018distributed,vepakomma2018split,osia2019hybrid,gu2019securing,hu2019fdml}. Their common idea is to let each data holder first use a partial neural network (i.e., encoder) to encode the raw input individually and then send the encoded representations to another data holder (or server) for the rest model training. 
Although such algorithmic methods are efficient, they have two shortcomings. 
First, the efficiency of those methods is usually at the expense of sacrificing model performance, as data holders train partial neural networks individually and the co-relation between data is not captured. 
Second, data privacy is not fully protected since the raw labels need to be shared with server during model training, as shown in Figure \ref{compare-exam-split}. Meanwhile the encoded representations may unintentionally reveal sensitive information (e.g., membership and property information~\cite{wu_sgld,collabrative_leakge,ganju2018property}).

\textbf{Cryptographic methods} focus on using pure cryptographic techniques, e.g., homomorphic encryption \cite{gilad2016cryptonets} or secure multi-party computation \cite{mohassel2017secureml,demmler2015aby}, for multi-parties to build privacy-preserving neural networks, as is shown in Figure \ref{compare-exam-sml}. 
Although such cryptographic methods have a strong privacy guarantee, they are difficult to scale to deep structures and large datasets due to the high communication and computational complexity of the cryptographic techniques. 
However, real-world applications always have two characteristics:
(1) datasets are large due to the massive data held by big companies; and
(2) high-performance neural network models are always deep and wide, which come with huge computational costs.
Therefore, efficiency becomes the main shortcoming when applying existing cryptographic methods in practice. 

\subsection{Our solution}

\nosection{Methodology}
In this paper, we propose a Scalable and Privacy-preserving deep Neural Network (\modelname) learning paradigm,
which combines the advantages of existing algorithmic methods and cryptographic methods. 

First, from the \textit{algorithmic perspective}, for scalability concern, we split the computation graph of a given DNN model into two types. 
The computations related to private data are performed by data holders and the rest heavy computations are delegated to a computation-efficient semi-honest server. 
Here, private data refers to the input and output of DNN models, i.e., features and labels. 
The hidden-layer-related computations in DNN involve many complicated non-linear operations, including active function such as Sigmoid and TanH and max-pooling, which are expensive for the cryptographical  techniques. 
Thus, by adopting a semi-honest server, these heavy computations could be done by the server in plaintext, similar as existing DNN models. 

Second, from the \textit{cryptographic perspective}, for accuracy and privacy concern, we let data holders securely calculate the first hidden layer collaboratively. 
More specifically, data holders firstly adopt cryptographic techniques, e.g., secret sharing and homomorphic encryption, to perform private feature related computations cooperatively. They generate the first hidden layer of DNN and send it to a semi-honest server. 
Then, the server performs the successive hidden layer related computations, gets the final hidden layer, and sends it to the data holder who has labels. 
The data holder who has labels conducts private label related computations and gets predictions based on the final hidden layer. 
The backward computations are performed reversely. 

In summary, private data and corresponding model are held by data holders, and the heavy non-private data related computations are done by the server. 
Our proposed \modelname~only involves cryptographic techniques for the first hidden layer, and therefore enjoys good scalability. 
To prevent privacy leakage of the hidden features on server, we further propose to inject moderate noises into the gradient during training. 
A typical way is to use differentially private stochastic gradient descent (DP-SGD). However, in practice it may lead to significant model accuracy drop~\cite{dp_sgd,pengFKGE2021}. 
In light of some recent works~\cite{wu_sgld}, we propose using Stochastic Gradient Langevin Dynamics (SGLD) to reduce the potential information leakage.

\nosection{Implementation}
We implement \modelname~in a decentralized network with three kinds of computation nodes, i.e., a coordinator, a server, and a group of clients. 
The coordinator splits the computation graph and controls the start and termination of \modelname~based on a certain condition, e.g., the number of iterations. 
The clients are data holders who are in charge of private data related computations, and the server is responsible for hidden layer related computations which can be adapted to existing deep learning backends such as PyTorch. 
Communications between workers and servers make sure the model parameters are correctly updated. 
Moreover, our implementation also supports user-friendly API, similar to PyTorch. Developers can easily build any privacy preserving deep neural network models without complicated cryptography knowledge. 

\nosection{Results}
We conduct experiments on real-world fraud detection and distress prediction datasets. Results demonstrate that our proposed \modelname~has comparable performance with the traditional neural networks that are trained on plaintext data. Moreover, experimental results also show that \modelname~significantly outperforms existing algorithmic methods and cryptographic approaches.

\nosection{Contributions} We summarize our main contributions as follows:
\begin{itemize} [leftmargin=*] \setlength{\itemsep}{-\itemsep}
\item We propose \modelname, a novel learning framework for scalable privacy preserving deep neural network, which not only has good scalability but also preserves data privacy. 
\item We implement \modelname~on decentralized network settings, which not only has user-friendly APIs but also can be adapted to existing deep learning backends such as PyTorch. 
\item Our proposal is verified on real-world datasets and the results show its superiority.
\end{itemize}


\section{Related Work}\label{background}
In this section, we briefly review two popular types of privacy preserving DNN models. 

\subsection{Algorithmic methods} 
These methods build privacy preserving DNN by split the computation graphs of DNN from algorithmic perspective \cite{gupta2018distributed,vepakomma2018split,osia2019hybrid,gu2019securing,hu2019fdml}. 
A common way is to let each data holder trains a partial neural network individually and then sends the hidden layers to another data holder (or server) for the rest model training \cite{gupta2018distributed}. 
For example, \cite{gu2019securing} proposed to enclose sensitive computation in a trusted execution environment, e.g., Intel Software Guard Extensions \cite{mckeen2013innovative}, to mitigate input information disclosures, and delegate non-sensitive workloads with hardware-assisted deep learning acceleration. 
\cite{vepakomma2018split} proposed splitNN, where each data holder trains a partial deep network model using its own features individually, and the partial models are concatenated 
followed by concatenating the partial models and sending to a server who has labels to train the rest of the model, as shown in Figure \ref{compare-exam-split}. 

However, 
the above-mentioned methods may suffer from accuracy and privacy problems, as we analyzed in Section \ref{sec-intro}. 
First, since data holders train partial neural networks individually and the correlation between the data held by different parties is not captured \cite{vepakomma2018split}, and therefore the accuracy is limited. 
Second, during model training of these methods, labels need to be shared with other participants such as the data holder or server \cite{gupta2018distributed,vepakomma2018split}, therefore, the data privacy are not fully protected. 

In this paper, our proposed \modelname~differs from existing algorithmic methods in mainly two aspects. 
First, we use cryptographic techniques for data holders to calculate the hidden layers collaboratively rather than compute them based on their plaintext data individually. 
By doing so, \modelname~can not only prevent a server from obtaining the individual information from each participant, but also capture feature interactions from the very beginning of the neural network, and therefore achieve better performance, as we will show in experiments. 
Second, \modelname~assumes both private feature and label data are held by participants themselves. Therefore, \modelname~can protect both feature and label privacy. 

\subsection{Cryptographic Methods}
Cryptographic methods are of two types, i.e., (1) customized methods for privacy preserving neural network, and (2) general frameworks that can be used for privacy preserving neural network. 

\subsubsection{Customized methods}
This type of methods designs specific protocols for privacy preserving neural networks using cryptographic techniques such as secure Multi-Party Computation (MPC) techniques and Homomorphic Encryption (HE). 
For example, some existing researches built privacy preserving neural network using HE \cite{yuan2013privacy,gilad2016cryptonets,hesamifard2017cryptodl,xu2019cryptonn}. To use these methods, the participants first need to encrypt their private data and then outsource these encrypted data to a server who trains neural networks using HE techniques. 
However, these methods have a drawback in nature. That is, they suffer from data abuse problem since the server can do whatever computations with these encrypted in hand. 
There are also researches built privacy preserving neural networks using MPC techniques such as secret sharing and garbled circuit \cite{juvekar2018gazelle,rouhani2018deepsecure,agrawal2019quotient,wagh2019securenn}. 

\subsubsection{General frameworks}
Besides the above customized privacy preserving neural network methods, there are also some general multi-party computation frameworks that can be used to build privacy preserving neural network models, e.g., SPDZ \cite{damgaard2012multiparty}, ABY \cite{demmler2015aby}, SecureML \cite{mohassel2017secureml}, ABY$^3$ \cite{mohassel2018aby}, and PrivPy \cite{li2018privpy}. 
Taking SecureML, a general privacy preserving machine learning framework, as an example, it provides three types of secret sharing protocols, i.e., Arithmetic sharing, Boolean sharing, and Yao sharing, and it also allows efficient conversions between plaintext and three types of sharing. 

However, all the above methods suffer from scalability problem. This is because deep neural networks contain many nonlinear active functions that are not cryptographically friendly. 
For example, existing works use polynomials \cite{chen2018logistic} or piece-wise function \cite{mohassel2017secureml,chen2019secure} to approximate continuous activation functions such as sigmoid and Tanh. And the piece-wise activation functions such as Relu \cite{glorot2011deep} rely on time-consuming secure comparison protocols. 
The polynomials and piece-wise functions not only reduce accuracy but more importantly, they significantly affect efficiency.
Therefore, these models are difficult to scale to deep networks and large datasets due to the high communication and computation complexities of the cryptographic techniques. 

In this paper, we propose \modelname~to combine the advantages of algorithmic methods and cryptographic methods. 
We use cryptographic techniques 
for data holders to calculate the first hidden layers securely, and delegate the heavy hidden layer related computations to a semi-honest server. 
Therefore, our proposal enjoys much better scalability comparing with the existing cryptographic methods, as we will report in experiments.

\section{Preliminaries}\label{pre}
In this section, we first briefly describe the data partition setting and threat model, 
and then present background knowledge on deep learning
, secret sharing
, and homomorphic encryption.

\subsection{Data Partition and Threat Model}\label{pre-setting}
\subsubsection{Data partition}
There are usually two types of data partition settings, i.e., \textit{horizontally data partitioning} and \textit{vertically data partitioning} in literature. 

The former one indicates each participant has a subset of the samples with the same features, while the latter denotes each party has the same samples but different features \cite{hall2011secure,yang2019federated}. 
In practice, the latter one is more common due to the fact that most users are always active on multi-platforms for different purposes, e.g., on Facebook for social and on Amazon for shopping. Therefore, we focus on vertically data partitioning in this paper. 

In practice, before building privacy preserving machine learning models under vertically data partitioning setting, the first step is to align samples between participants, e.g., align users when each sample is user features and lable. 
Taking a fraud user detection scenario for example, assume two companies both have a batch of users with different user features, and they want to build a better fraud detection system collaboratively and securely. To train a fraud detection model such as neural network, they need match the intersectant users and align them as training samples. 
This can be done efficiently using the existing \textit{private set intersection} technique \cite{de2010practical,pinkas2014faster}. In this paper, we assume participants have already aligned samples and are ready for building privacy preserving neural network. 

\subsubsection{Threat Model}
There are two commonly used threat models in literature, i.e., \textit{semi-honest} (passive) adversary and \textit{malicious} (active) adversary are two commonly used threat model in literature \cite{mohassel2017secureml,li2018privpy}. 
The former assumes that participants will strictly follow the protocol, but they try to infer as much information as possible using all intermediate computation results. The latter accepted the situation where the corrupt participants can arbitrarily deviate from the protocol specification. 
Although the semi-honest model may be “compiled” into a protocol secure against malicious adversaries, but they are usually too inefficient for practical use \cite{goldreich2009foundations}. Therefore, we aim to build privacy preserving neural network model under semi-honest adversary setting. That is, we assume all participants (data holders) and the server will strictly follow our proposed protocol. We also assume that participants will not collude with the server. 

\subsection{Deep Neural Network}\label{pre-nn}
Deep Neural Network (DNN) has been showing great power in kinds of machine learning tasks, since it can learn complex functions by composing multiple non-linear modules to transform representations from low-level raw inputs to high-level abstractions \cite{gu2019securing}. 
Mathematically, the forward procedure of a DNN can be defined as a representation function $f$ that maps an input $\textbf{X}$ to an output $\hat{y}$, i.e., $\hat{y}=f(\textbf{X}, \bm\theta)$, where $\bm\theta$ is model parameter. 
Assume a DNN has $L$ layers, then $f$ is composed of $L$ sub-functions $f_{l|{l\in[1,L]}}$, which are connected in a chain. That is, $f(\textbf{X})=f_L(...f_2(f_1(\textbf{X}, \bm\theta_0), \bm\theta_1)..., \bm\theta_{L-1})$, as is shown in Figure \ref{compare-exam-nn}.

\subsubsection{Model learning}
The above model parameter can be learnt by using mini-batch gradient descent. 
Let $\mathcal{D}=\{(\textbf{x}_i, y_i)\}_{i=1}^n$ be the training dataset, where $n$ is the sample size, $\textbf{x}_i$ is the feature of $i$-th sample, and $y_i$ is its corresponding label. 
The loss function of DNN is built by minimizing the losses over all the training samples, that is, $\mathcal{L}=\sum_i^n l(y_i, \hat{y_i})$.  Here $l(y_i, \hat{y_i})$ is defined based on different tasks, e.g., softmax for classification tasks. 
After it, DNN can be learnt efficiently by minimizing the losses using mini-batch Stochastic Gradient Descent (SGD) and its variants. 
Take mini-batch gradient descent for example, let $\textbf{B}$ be samples in each batch, $|\textbf{B}|$ be the batch size, $\textbf{X}_B$ and $\textbf{Y}_B$ be the features and labels in the current batch, then the model of DNN can be updated by: 
\begin{equation}\label{batch-update}
\boldsymbol{\theta} \leftarrow \boldsymbol{\theta} - \frac{\alpha}{|\textbf{B}|} \cdot \frac{\partial \mathcal{L}}{\partial \boldsymbol{\theta}} ,
\end{equation}
where $\alpha$ is the learning rate.
The model gradient $\frac{\partial \mathcal{L}}{\partial \boldsymbol{\theta}}$ is usually calculated by back propagation \cite{goodfellow2016deep}.
As discussed in the introduction, the hidden features can impose some sensitive information. To reduce the information lekage, in this paper, we propose to use SGLD~\cite{sgld_original}, a Bayesian learning approach. Specifically, SGLD can be seen as a noisy version of the conventional SGD algorithm. To reduce the leakage, SGLD injects an isotropic
Gaussian noise vector into the gradients. Formally, this process can be represented as:
\begin{equation}
    \boldsymbol{\theta} \leftarrow \boldsymbol{\theta} - (\frac{\alpha_t}{2}\frac{\partial \mathcal{L}}{\partial \boldsymbol{\theta}}+\eta_t), \eta_t\sim\mathcal{N}(0, \alpha_t \mathbf{I}),
\label{eq:sgld}
\end{equation}
here $\alpha_t$ denotes the learning rate at the $t$-th iteration and $\mathcal{N}(0,\alpha_t\mathbf{I})$ is the Gaussian distribution.

\subsection{Arithmetic Secret Sharing}\label{pre-ss}

Assume there are two parties ($\mathcal{P}_0$ and $\mathcal{P}_1$),  $\mathcal{P}_0$ has an $\ell$-bit secret $a$ and $\mathcal{P}_1$ has an $\ell$-bit secret $b$. 
To secretly share $\textbf{Shr}(\cdot)$ $a$ for $\mathcal{P}_0$, party $\mathcal{P}_0$ generates an integer $r \in \mathds{Z}_{2^\ell}$ uniformly at random, sends $r$ to party $\mathcal{P}_1$ as a share $\left\langle a \right\rangle_1$, and keeps $\left\langle a \right\rangle_0 = a- r$ mod $2^\ell$ as the other share. $\mathcal{P}_1$ can share $b$ with $\mathcal{P}_0$ similarly, and $\mathcal{P}_1$ keeps $\left\langle b \right\rangle_1$ and $\mathcal{P}_0$ receives $\left\langle b \right\rangle_0$. 
We will describe how to perform addition and multiplication and how to support decimal numbers and vectors in the following subsections. 

\subsubsection{Addition and Multiplication}
Suppose $\mathcal{P}_0$ and $\mathcal{P}_1$ want to secretly calculate $a+b$ using Arithmetic sharing, $\mathcal{P}_0$ locally calculates $\left\langle c \right\rangle_0=\left\langle a \right\rangle_0 + \left\langle b \right\rangle_0$ mod $2^\ell$ and $\mathcal{P}_1$ locally calculates $\left\langle c \right\rangle_1=\left\langle a \right\rangle_1 + \left\langle b \right\rangle_1$ mod $2^\ell$. 
To reconstruct $\textbf{Rec}(\cdot, \cdot)$ a secret, one party sends his share to the other party who reconstruct the plaintext by $c=\left\langle c \right\rangle_0 + \left\langle c \right\rangle_1$ which is equal to $a+b$.
To secretly calculate $a \cdot b$ using Arithmetic sharing, Beaver’s multiplication triples \cite{beaver1991efficient} are usually involved. 
Specifically, to multiply two secretly shared values ($a$ and $b$), $\mathcal{P}_0$ and $\mathcal{P}_1$ first need to collaboratively generate a triple $\langle u \rangle$, $\langle v \rangle$, and $\langle w \rangle$, where $u, v$ are uniformly random values in $\mathds{Z}_{2^\ell}$ and $w=u \cdot v$ mod $2^\ell$. 
Then, $\mathcal{P}_0$ locally computes $\langle e \rangle _0 = \langle a \rangle _0 - \langle u \rangle _0$ and $\langle f \rangle _0 = \langle b \rangle _0 - \langle v \rangle _0$, and $\mathcal{P}_1$ locally computes $\langle e \rangle _1 = \langle a \rangle _1 - \langle u \rangle _1$ and $\langle f \rangle _1 = \langle b \rangle _1 - \langle v \rangle _1$. 
Next, they reconstruct $e$ and $f$ by $\textbf{Rec}(\langle e \rangle _0, \langle e \rangle _1)$ and $\textbf{Rec}(\langle f \rangle _0, \langle f \rangle _1)$, respectively. 
Finally, $\mathcal{P}_0$ gets $\langle c \rangle _0 = f \cdot \langle a \rangle _0 + e \cdot \langle b \rangle _0 + \langle w \rangle _0$ and $\mathcal{P}_1$ gets $\langle c \rangle _0 = e \cdot f + f \cdot \langle a \rangle _1 + e \cdot \langle b \rangle _1 + \langle w \rangle _1$, where $\langle c \rangle _0 + \langle c \rangle _1 = a \cdot b$. 

\subsubsection{Supporting decimal numbers and vectors}
The above protocols only work in finite field, since it needs to sample uniformly in $\mathds{Z}_{2^\ell}$. However, in neural network, both features and model parameters are usually decimal vectors. 
To support this, we adopt the existing fixed-point representation to approximate decimal arithmetics efficiently \cite{mohassel2017secureml}. 
Simply speaking, we use at most $l_F$ bits to represent the fractional part of decimal numbers. 
Specifically, Suppose $a$ and $b$ are two decimal numbers with at most $l_F$ bits in the fractional part, to do fixed-point multiplication, we first transform them to integers by letting $a'=2^{l_F}a$ and $b'=2^{l_F}b$, and then calculate $c=a'b'$. Finally, we truncate the last $l_F$ bits of $c$ so that it has at most $l_F$ bits representing the fractional part, with $l_F=16$ in this paper. It has been proven that this truncation technique also works when $c$ is secret shared \cite{mohassel2017secureml}. 
We set $l_F=16$ in this paper. 
After this, it is easy to vectorize the addition and multiplication protocols under Arithmetic sharing setting. 
We will present how participants use arithmetic sharing to calculate the first hidden layer cooperatively in Section \ref{model-feature}. 

\subsection{Additive Homomorphic Encryption}\label{pre-he}

Additive Homomorphic Encryption (HE) is a kind of encryption scheme which allows a third party (e.g., cloud, service provider) to perform addition on the encrypted data while preserving the features of addition operation
and format of the encrypted data \cite{acar2018survey}.  
Suppose there is a server with key generation ability and a number of participants with private data. Under such setting, the use of additive HE mainly has the following steps \cite{acar2018survey}:
\begin{itemize}[leftmargin=*] \setlength{\itemsep}{-\itemsep}
    \item \textbf{Key generation. } The server generates the public and secret key pair $(pk, sk)$ and distributes public key $pk$ to the participants.
    \item \textbf{Encryption. } Given a plaintext $x$ on any participant, it is encrypted using $pk$ and a random $r$, i.e., $\llbracket x \rrbracket=\textbf{Enc}(pk; x, r)$, where $\llbracket x \rrbracket$ denotes the ciphertext and $r$ makes sure the ciphertexts are different in multiple encryptions even with the same plaintexts. 
    \item \textbf{Homomorphic addition. } Given two ciphertext ($\llbracket x \rrbracket$ and $\llbracket y \rrbracket$) on participants, addition can be done by $\llbracket x+y \rrbracket=\llbracket x \rrbracket+\llbracket y \rrbracket$.
    \item \textbf{Decryption. } Given a ciphertext $\llbracket x \rrbracket$ on server, it can be decrypted by $x=\textbf{Dec}(sk; \llbracket x \rrbracket)$. 
\end{itemize}

In this paper, we choose Paillier \cite{paillier1999public} to do additive HE, which is popularly used due to its high efficiency. We will present how participants use additive HE to calculate the first hidden layer cooperatively in Section \ref{model-feature}. 
\section{The Proposed Method}\label{model}
In this section, we first describe the problem 
, and then present the overview of \modelname
. 
Next, we present the sub-modules of \modelname~in details
, and finally present the learning of \modelname.


\subsection{Problem Description}\label{model-problem}
We start from a concrete example. 
Suppose there are two financial companies, i.e., $\mathcal{A}$ and $\mathcal{B}$, who both need to detect fraud users. 
As is shown in Figure \ref{framework}, 
$\mathcal{A}$ has some user features ($\textbf{X}_A$, shown in yellow dots) and labels ($\textbf{y}$, shown in yellow squares), and $\mathcal{B}$ has features ($\textbf{X}_B$, shown in green dots) for the same batch of users. 
Although $\mathcal{A}$ can build a Deep Neural Network (DNN) for fraud detection using its own data, the model performance can be improved by incorporating features of $\mathcal{B}$. 
However, these two companies can not share data with each other due to the fact that leaking users' private data is against regulations. 
This is a classic data isolation problem. 
It is challenging for both parties to build scalable privacy preserving neural networks collaboratively without compromising their private data. 
In this paper, we only consider the situation where two data holders have the same sample set, one of them ($\mathcal{A}$) has partial features and labels, and the other ($\mathcal{B}$) has the rest partial features. 
Our proposal can be naturally extended to more than two parties. 

\subsection{Proposal Overview}\label{model-overview}
We propose a novel scalable and privacy-preserving deep neural network learning framework (\modelname) for the above challenge. 
As described in Section \ref{pre-nn}, DNN can be defined as a layer-wise representation function. 
Motivated by the existing work \cite{gupta2018distributed,vepakomma2018split,osia2019hybrid,gu2019securing}, we propose to decouple the computation graph of DNN into two types, i.e., the computations related to private data are performed by data holders using cryptographic techniques, and the rest computations are delegated to a semi-honest server with high computation ability. 
Here, the private data are the input and output of the neural network, which corresponds to the private features and labels from data holders. 
By letting the semi-honest server perform the heavy computations in the hidden layers of DNN in plaintext, we can avoid many complicated non-linear operations, e.g., non-linear active function and max-pooling, which are expensive for the cryptographical  techniques. 
The extra semi-honest is also the key to scalable privacy-preserving DNN. 

\begin{figure}[t]
\centering
\includegraphics[width=8cm]{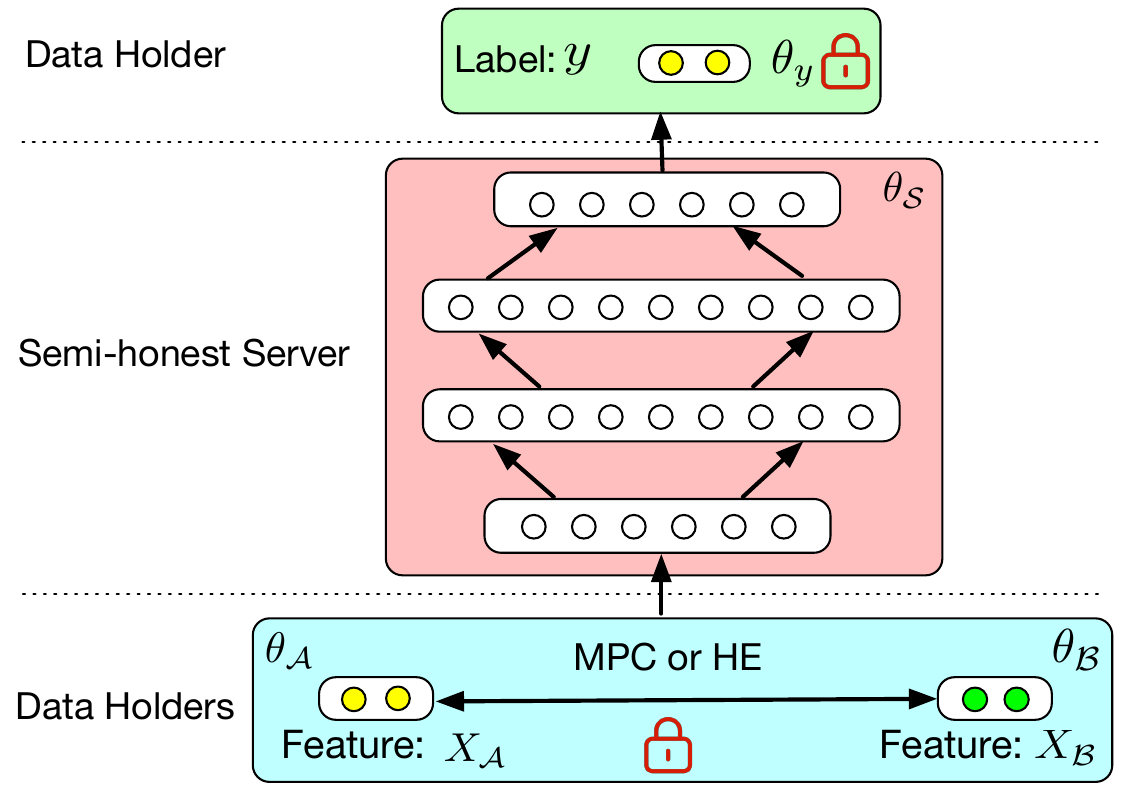}
\caption{The proposed \modelname, which is trained by using SGLD. The bottom blue part stands for private feature related computations which
are performed by data holders collaboratively and securely (Section \ref{model-feature}). 
The middle pink part is hidden layer related computations that are conducted on by a semi-honest server (Section \ref{model-hidden}). 
The top green part is private label related computations that are done by the data holder who has label (Section \ref{model-label}).}
\label{framework}
\end{figure}

Specifically, we divide the model parameters ($\theta$) into three parts, (1) \textit{the computations that are related to private features on both data holders} ($\theta_\mathcal{A}$ and $\theta_\mathcal{B}$), (2) \textit{the rest heavy hidden layer related computations on server} ($\theta_\mathcal{S}$), and (3)  \textit{the computations related to private labels on the data holder who has label} ($\theta_y$). 
As shown in Figure \ref{framework}, the first part is private data related computations and therefore are performed by data holders themselves using secret sharing and homomorphic encryption techniques, the second part is delegated to a semi-honest server which has rich computation resources. 
We summarize the forward propagation in Algorithm \ref{algo}.
We will describe each part in details in the following subsections. 

\begin{algorithm}[t]
\caption{The forward propagation of \modelname}\label{algo}
\KwIn {Features of $\mathcal{A}$ ($\textbf{X}_A$), features of $\mathcal{B}$ ($\textbf{X}_B$), server ($\mathcal{S}$), and the number of iteration ($T$)}
\KwOut{Predictions ($\hat{\textbf{y}}$) on $\mathcal{A}$}

$\mathcal{A}$ initializes $\bm{\theta}_A$ and $\bm{\theta}_y$, $\mathcal{B}$ initializes $\bm{\theta}_B$, and the server initializes $\bm{\theta}_S$ \\

\For{$t=1$ to $T$}
{
	\For{each mini-batch in training datasets}
	{
		\# private feature related computations by $\mathcal{A}$ and $\mathcal{B}$ (Section \ref{model-feature}) \\
		$\mathcal{A}$ and $\mathcal{B}$ collaboratively learn the first hidden layer ($\textbf{h}_1$) using secret sharing (Algorithm \ref{model-feature-ss}) or homomorphic encryption (Algorithm \ref{model-feature-he}), i.e.,  $\textbf{h}_1 = f(\textbf{X}_A, \textbf{X}_B; \bm{\theta}_A, \bm{\theta}_B)$ \\
		\# hidden layer related computations by Server (Section \ref{model-hidden}) \\
		$\mathcal{S}$ calculates the final hidden layer by $\textbf{h}_L = f(\textbf{h}_1; \bm{\theta}_S)$ \\
		\# private label related computations by $\mathcal{A}$ (Section \ref{model-label}) \\
		$\mathcal{A}$ makes prediction by $\hat{\textbf{y}} = f(\textbf{h}_L; \bm{\theta}_y)$
		
	}
}
\Return Predictions ($\hat{\textbf{y}}$) on $\mathcal{A}$
\end{algorithm}

\begin{algorithm}[t]
\caption{Data holders $\mathcal{A}$ and $\mathcal{B}$ securely calculate the first hidden layer using arithmetic sharing}\label{model-feature-ss}
\KwIn {features of $\mathcal{A}$ and $\mathcal{B}$ ($\textbf{X}_A$ and $\textbf{X}_B)$, current models of $\mathcal{A}$ and $\mathcal{B}$ ($\bm{\theta}_A$ and $\bm{\theta}_B)$, and Server ($\mathcal{S}$)}
\KwOut{The first hidden layer $\textbf{h}_1$ on $\mathcal{S}$}

$\mathcal{A}$ and $\mathcal{B}$ locally generate $\left\langle\textbf{X}_A\right\rangle_1$ and $\left\langle\textbf{X}_A\right\rangle_2$, and $\left\langle\textbf{X}_B\right\rangle_1$ and $\left\langle\textbf{X}_B\right\rangle_2$, respectively \label{algo-ss-1}\\
$\mathcal{A}$ and $\mathcal{B}$ locally generate $\left\langle\bm{\theta}_A\right\rangle_1$ and $\left\langle\bm{\theta}_A\right\rangle_2$, and $\left\langle\bm{\theta}_B\right\rangle_1$ and $\left\langle\bm{\theta}_B\right\rangle_2$, respectively\\
$\mathcal{A}$ distributes $\left\langle\textbf{X}_A\right\rangle_2$ and $\left\langle\bm{\theta}_A\right\rangle_2$ to $\mathcal{B}$ \\
$\mathcal{B}$ distributes $\left\langle\textbf{X}_B\right\rangle_1$ and $\left\langle\bm{\theta}_B\right\rangle_1$ to $\mathcal{A}$ \label{algo-ss-4}\\
$\mathcal{A}$ locally calculates $\left\langle\textbf{X}\right\rangle_1 = \left\langle\textbf{X}_A\right\rangle_1 \oplus \left\langle\textbf{X}_B\right\rangle_1$, $\left\langle\bm{\theta}\right\rangle_1 = \left\langle\bm{\theta}_A\right\rangle_1 \oplus \left\langle\bm{\theta}_B\right\rangle_1$, and $\left\langle\textbf{X}\right\rangle_1 \cdot \left\langle\bm{\theta}\right\rangle_1$ \label{algo-ss-ct1}\\
$\mathcal{B}$ locally calculates $\left\langle\textbf{X}\right\rangle_2 = \left\langle\textbf{X}_A\right\rangle_2 \oplus \left\langle\textbf{X}_B\right\rangle_2$, $\left\langle\bm{\theta}\right\rangle_2 = \left\langle\bm{\theta}_A\right\rangle_2 \oplus \left\langle\bm{\theta}_B\right\rangle_2$, and $\left\langle\textbf{X}\right\rangle_2 \cdot \left\langle\bm{\theta}\right\rangle_2$ \label{algo-ss-ct2}\\

$\mathcal{A}$ and $\mathcal{B}$ calculate $\left\langle\textbf{X}\right\rangle_1 \cdot \left\langle\bm{\theta}\right\rangle_2$ and $\left\langle\textbf{X}\right\rangle_2 \cdot \left\langle\bm{\theta}\right\rangle_1$ using arithmetic sharing matrix multiplication, $\mathcal{A}$ get $\left\langle\textbf{X}_1 \cdot \bm{\theta}_2\right\rangle_A $ and $\left\langle\textbf{X}_2 \cdot \bm{\theta}_1\right\rangle_A $, $\mathcal{B}$ gets $\left\langle\textbf{X}_1 \cdot \bm{\theta}_2\right\rangle_B $ and $\left\langle\textbf{X}_2 \cdot \bm{\theta}_1\right\rangle_B $ \label{algo-ss-smm}\\ 

$\mathcal{A}$ locally calculates $\left\langle\textbf{X} \cdot \bm{\theta}\right\rangle_A = \left\langle\textbf{X}\right\rangle_1 \cdot \left\langle\bm{\theta}\right\rangle_1 + \left\langle\textbf{X}_1 \cdot \bm{\theta}_2\right\rangle_A  + \left\langle\textbf{X}_2 \cdot \bm{\theta}_1\right\rangle_A$  \label{algo-ss-rc1}\\

$\mathcal{B}$ locally calculates $\left\langle\textbf{X} \cdot \bm{\theta}\right\rangle_B = \left\langle\textbf{X}\right\rangle_2 \cdot \left\langle\bm{\theta}\right\rangle_2 + \left\langle\textbf{X}_1 \cdot \bm{\theta}_2\right\rangle_B  + \left\langle\textbf{X}_2 \cdot \bm{\theta}_1\right\rangle_B$ \label{algo-ss-rc2}\\

$\mathcal{A}$ and $\mathcal{B}$ sends $\left\langle\textbf{X} \cdot \bm{\theta}\right\rangle_A$ and $\left\langle\textbf{X} \cdot \bm{\theta}\right\rangle_B$ to $\mathcal{S}$

$\mathcal{S}$ calculates $\textbf{h}_1=\left\langle\textbf{X} \cdot \bm{\theta}\right\rangle_A + \left\langle\textbf{X} \cdot \bm{\theta}\right\rangle_B$
\end{algorithm}

\subsection{Private Feature Related Computations}\label{model-feature}

Private feature related computations refer to data holders collaboratively calculate the hidden layer of a DNN using their private features. 
Here, data holders want to (1) calculate a common function, i.e., $\textbf{h}_1 = f(\textbf{X}_A, \textbf{X}_B; \bm{\theta}_A, \bm{\theta}_B)$, collaboratively and (2) keep their features, i.e., $\textbf{X}_A$ and $\textbf{X}_B$, private. 
Mathematically, $\mathcal{A}$ and $\mathcal{B}$ have partial features ($\textbf{X}_A$ and $\textbf{X}_B$) and partial model parameters ($\bm{\theta}_A$ and $\bm{\theta}_B$), respectively, and they want to compute the output of the first hidden layer collaboratively. 
That is, $\mathcal{A}$ and $\mathcal{B}$ want to compute 
$\textbf{h}_1 = \textbf{X}_A \cdot \bm{\theta}_A + \textbf{X}_B \cdot \bm{\theta}_B = (\textbf{X}_A \oplus \textbf{X}_B) \cdot (\bm{\theta}_A \oplus \bm{\theta}_B)$, 
where $\oplus$ denotes concatenation operation. 
Note that we omit the activation function here, since activation can be done by server after it receives $\textbf{h}_1$ from data holders. 

The above secure computation problem can be done by cryptographical techniques. 
As we described in Section \ref{pre}, arithmetic sharing and additive HE are popularly used due to their high efficiency. We will present two solutions based on arithmetic sharing and additive HE, respectively. 

\subsubsection{Arithmetic sharing based solution}
We first present how to solve the above secure computation problem using arithmetic sharing. 
The main technique is secret sharing based matrix addition and multiplication on fixed-point numbers, please refer to Section~\ref{pre-ss} for more details. 
We summarize the secure protocol in Algorithm \ref{model-feature-ss}. 
As Algorithm~\ref{model-feature-ss} shows, data holders first secretly \textit{share} their feature and model (Lines \ref{algo-ss-1}-\ref{algo-ss-4}), then they concat the feature and model shares \ref{algo-ss-ct1}-\ref{algo-ss-ct2}. 
After it, data holders calculate $\textbf{h}_1= \textbf{X} \cdot \bm{\theta}$ by using distributive property, i.e., $\textbf{X} \cdot \bm{\theta} = (\left\langle\textbf{X}\right\rangle_1 + \left\langle\textbf{X}\right\rangle_2) \cdot (\left\langle\bm{\theta}\right\rangle_1 + \left\langle\bm{\theta}\right\rangle_2)$ (Line \ref{algo-ss-smm}). 
Next, each data holder sums up their intermediate shares as shares of the hidden layer (Lines \ref{algo-ss-rc1}-\ref{algo-ss-rc2}).  
To this end, $\mathcal{A}$ and $\mathcal{B}$ each obtains a partial share of the hidden layer, i.e., $\langle \textbf{h}_1 \rangle _A = \left\langle\textbf{X} \cdot \bm{\theta}\right\rangle_A$ and $\langle \textbf{h}_1 \rangle _B = \left\langle\textbf{X} \cdot \bm{\theta}\right\rangle_B$. 
Finally, the server \textit{reconstruct} the first hidden layer by 
$\textbf{h}_1 = \langle \textbf{h}_1 \rangle _A + \langle \textbf{h}_1 \rangle _B$. 

\subsubsection{Additive HE based solution}
We then present how to solve the secure computation problem using additive HE. We summarize the protocol in Algorithm \ref{model-feature-he}, where we first rely on a semi-honest server to generate key pair and decryption (Line \ref{algo-he-keygen}), then let data holders calculate the encrypted hidden layer (Lines \ref{algo-he-a}-\ref{algo-he-b}), and finally let the server decrypt to get the plaintext hidden layer (Line \ref{algo-he-s}). 

Arithmetic sharing and additive HE have their own advantages. 
Arithmetic sharing does not need time-consuming encryption and decryption operations, however, it has higher communication complexity. 
In contrast, although additive HE has lower communication complexity, it relied on the time-consuming encryption and decryption operations. We will empirically study their performance under different network settings in Section \ref{sec-exp-speed}. 

\begin{algorithm}[t]
\caption{Data holders $\mathcal{A}$ and $\mathcal{B}$ securely calculate the first hidden layer using additive HE}\label{model-feature-he}
\KwIn {features of $\mathcal{A}$ and $\mathcal{B}$ ($\textbf{X}_A$ and $\textbf{X}_B)$, current models of $\mathcal{A}$ and $\mathcal{B}$ ($\bm{\theta}_A$ and $\bm{\theta}_B)$, and Server ($\mathcal{S}$)}
\KwOut{The first hidden layer $\textbf{h}_1$ on $\mathcal{S}$}

$\mathcal{S}$ generates key pair $(pk, sk)$ and distributes public key $pk$ to $\mathcal{A}$ and $\mathcal{B}$ \label{algo-he-keygen} \\
$\mathcal{A}$ calculates $\textbf{X}_A \cdot \bm{\theta}_A$, encrypts it with $pk$, and sends $\llbracket \textbf{X}_A \cdot \bm{\theta}_A \rrbracket$ to $\mathcal{B}$ \label{algo-he-a} \\
$\mathcal{B}$ calculates $\textbf{X}_B \cdot \bm{\theta}_B$, encrypts it with $pk$, calculates $\llbracket \textbf{X}_A \cdot \bm{\theta}_A + \textbf{X}_B \cdot \bm{\theta}_B \rrbracket = \llbracket \textbf{X}_A \cdot \bm{\theta}_A \rrbracket + \llbracket \textbf{X}_B \cdot \bm{\theta}_B \rrbracket$, and sends it to $\mathcal{S}$ \label{algo-he-b} \\
$\mathcal{S}$ decrypt $\llbracket \textbf{X}_A \cdot \bm{\theta}_A + \textbf{X}_B \cdot \bm{\theta}_B \rrbracket$ using $sk$ and gets $\textbf{h}_1 = \textbf{X}_A \cdot \bm{\theta}_A + \textbf{X}_B \cdot \bm{\theta}_B$  \label{algo-he-s}

\Return $\textbf{h}_1$ on $\mathcal{S}$
\end{algorithm}

\subsection{Hidden Layer Related Computations}\label{model-hidden}
After $\mathcal{A}$ and $\mathcal{B}$ obtain the shares of the first hidden layer, they send them to a semi-honest server for hidden layer related computations, i.e., $\textbf{h}_{L} = f (\textbf{h}_1, \bm{\theta}_{S})$. 
This is the same as the traditional neural networks. 
Given $l$-th hidden layer $\textbf{h}_l$, where $1 \le l \le L-1$ and $L$ be the number of hidden layers, the $(l+1)$-th hidden layer can be calculated by 
\begin{equation}\label{hdl}
\textbf{h}_{l+1} = f_l (\textbf{h}_l, \bm{\theta}_{l}),
\end{equation}
where $\bm{\theta}_{l}$ is the parameters in $l$-th layer, and $f_l$ is the active function of the $l$-th layer. 
These are the most time-consuming computations, because there are many nonlinear operations, e.g., max pooling, are not cryptographically
friendly. We leave these heavy computations on a server who has strong computation power. 
To this end, our model can scale to large datasets. 

Moreover, one can easily implement any kinds of deep neural network models using the existing deep learning platforms such as TensorFlow (https://tensorflow.org/) 
and PyTorch (https://pytorch.org/). 
As a comparison, for the existing privacy preserving neural network approaches such as SecureML \cite{mohassel2017secureml} and ABY \cite{demmler2015aby}, one needs to design specific protocols for different deep neural network models, which significantly increases the development cost for deep learning practitioners. 

\subsection{Private Label Related Computations}\label{model-label}
After the neural server finishes the hidden layer related computations, it sends the final hidden layer $\textbf{h}_L$ to the data holder who has the label, i.e., $\mathcal{A}$ in this case, for computing predictions. That is 
\begin{equation}\label{eq-predict}
\hat{\textbf{y}} = \delta (\textbf{h}_L, \bm{\theta}_y),
\end{equation}
where $\delta$ is designed based on different prediction tasks, e.g., $\delta$ is the softmax function for classification tasks. 

\subsection{Learning Model Parameters}\label{model-learn}
Note that the security assumption our \modelname~is that the semi-honest server cannot infer the private information of the data holders from the hidden layer of DNN. 
Adversary learning and Bayesian deep learning are two effective ways to prevent potential information leakage from the hidden layers of DNN. 
The former lets each data holder train a defender model who tries to simulate the behaviors of an attacker, i.e., the defender tries to infer the private data given the output of the hidden layer \cite{osia2019hybrid}. 
The later aims to learn the distribution of a DNN model rather than the specific model weight, and the representative work is Stochastic Gradient Langevin Dynamics (SGLD)~\cite{wu_sgld}. 
Specifically, in this paper, to prevent the information leakage caused by the hidden features, i.e., ${\mathbf{h}_l}_{l\ge 1}$, we propose using SGLD instead of SGD to optimize the parameters of \modelname. The formal description of SGLD is introduced in Equation~\ref{eq:sgld} and we refer reader to the prior work~\cite{sgld_original} for more details.

The gradient is computed using back propagation following the chain-rule, which is similar to the forward propagation procedure in Algorithm \ref{algo}. 
Specifically, given the loss calculated by the data holder who has label, this data holder first calculates the gradients of its model parameters, and then sends the model update to the server. The server then calculates the gradient of its own model parameters layer-by-layer, and then sends the model update of the first several hidden layers (i.e., the private feature related layers) to data holders. 
Finally, each data holder calculates the corresponding model gradient. 
Both forward computation and backward computation need communication between $\mathcal{A}$, $\mathcal{B}$, and the server, in a decentralized manner. 
During training, all the private data ($\textbf{X}_A$, $\textbf{X}_B$, and $\textbf{y}$) and private data related model parameters ($\bm{\theta}_A$, $\bm{\theta}_B$, and $\bm{\theta}_y$) are kept by data holders. Therefore, data privacy is preserved to a large extent. 

It is worth noticing that our proposal can be generalized to multi-parties and the situations that the data holders collaboratively calculate $i$ ($1 \le i \le L$) hidden layers instead of the first hidden layer only. Therefore, the existing method \cite{mohassel2017secureml} is one of our special cases, i.e., $\mathcal{A}$ and $\mathcal{B}$ collaboratively calculate all the neural networks using cryptographical techniques, without the semi-honest server.

\section{Implementation}\label{imple}
In this section, we present the implementation of \modelname~and showcase its user-friendly APIs by an example. 

\begin{figure}[t]
\centering
\includegraphics[width=5cm]{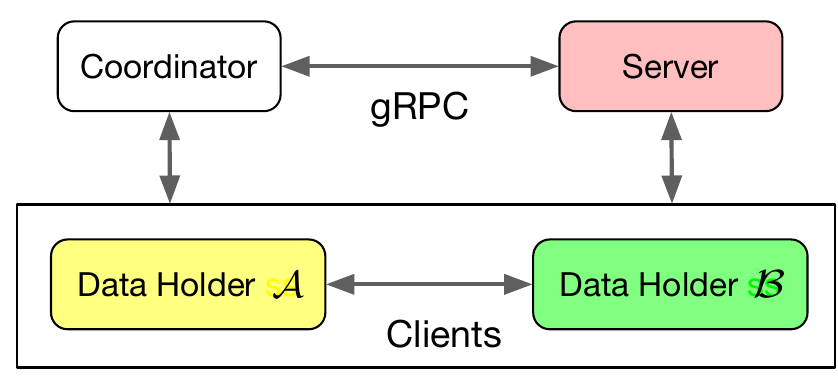}
\caption{Implementation framework of \modelname.}
\label{imple-framework}
\end{figure}

\subsection{Overview}\label{imple-overview}
We implement \modelname~in a decentralized network, where there are three kinds of computation nodes, i.e., a coordinator, a server, and a group of clients, as is shown in Figure \ref{imple-framework}. 
\textit{The coordinator} controls the start and terminal of \modelname. 
When \modelname~starts, the coordinator split the computation graph into three parts, sends each part to the corresponding clients and server, and notifies clients and server to begin training/prediction. 
Meanwhile, the coordinator monitors the status and terminates the program if it reaches a certain pre-defined condition, e.g., the number of iterations. 
Note that, during the model training and prediction procedures, the coordinator only controls the status of the program but cannot directly or indirectly touch any private input and intermediate result. 
In our implementation, we use a standalone machine to act as the role of the coordinator who gives orders to the server and data holders. Since our threat model is semi-honest in our paper, we assume the coordinator will strictly follow the protocol and give the right orders to the participants. 
\textit{The clients} are data holders who are in charge of private data related computations, and \textit{the server} is responsible for hidden layer related computations which can be adapted to existing deep learning backends such as PyTorch. We will describe the details of data holders and server below. 

\subsection{Implementation Details}\label{imple-detail}
The detailed implementation mainly includes the computations on data holders, the computations on server, and communications. 

\subsubsection{Computations on data holders}
We implement the forward and backward computations by clients (data holders) using Python and PyTorch. 
First, for the private feature related computations by clients collaboratively,  when clients receive orders from the coordinator, they first initialize model parameters and load their own private features, and then make calculations  following Algorithm \ref{model-feature-ss} and Algorithm \ref{model-feature-he}. 
We implement the private feature related computations using Python. 
Second, for the private label related computations by the client who has label, when the client receives the last hidden layer from the server, it initializes the model parameter and makes prediction based on Eq. \eqref{eq-predict}. 
The private label related computations are done in PyTorch automatically.

\subsubsection{Computations on server}
For the heavy hidden layer related computations on server, we also use PyTorch as backend to perform the forward and backward computations. 
Specifically, after server receives the first hidden layer from clients, it takes the first hidden layer as the input of a PyTorch network structure to make the hidden layer related computations, and gets the last hidden layer on server, and then the client who has label makes predictions. 
Both the forward and backward computations are made automatically on PyTorch. 
Note that private label related computations on client and the heavy hidden layer related computations on server are done using the ``model parallel'' mechanism in PyTorch \cite{paszke2019pytorch}. 

\subsubsection{Communications} 
Communications between the coordinator, server, and clients make sure the model parameters are correctly updated.
We adopt Google's gRPC\footnote{https://grpc.io/} as the communication protocol. 
Before training/prediction, we configure detailed parameters for clients and on the coordinator, such as the IP addresses, gateways, and dataset locations. 
At the beginning of the training/prediction, the coordinator shakes hands with clients and server to build connection. 
After that, they exchange data to finish model training/prediction as described above.

\subsection{User-friendly APIs}\label{imple-api}

Our implementation supports user-friendly API, similar as PyTorch. Developers can easily build any privacy preserving deep neural network models without the complicated cryptography knowledge. 
Figure \ref{examcode} shows an example code of \modelname, which is a neural network with network structure $(64, 256, 512, 256, 64, 2)$. Here, we assume two clients (A and B) each has 32-dimensional input features, and A has the 5-classes labels. 
From Figure \ref{examcode}, we can find that the use of \modelname~is quite the same as PyTorch, and the most different steps are the forward and backward computations of the first hidden layer by clients using cryptographic techniques (Line 35 and Line 44). 


\begin{figure}[t]
\centering
\begin{lstlisting}[basicstyle=\ttfamily\footnotesize, language=cement]
import torch
import torch.nn as nn
import torch.optim as optim

class ToyModel(nn.Module):
    def __init__(self):
        super(ToyModel, self).__init__()
        
        # server makes hidden layer related computations
        self.hidden1 = torch.nn.ReLU(256, 512).to('server')
        self.hidden2 = torch.nn.sigmoid(512, 256).to('server')
        self.hidden3 = torch.nn.sigmoid(256, 64).to('server')
        
        # A makes private label related computations
        self.output = torch.nn.ReLU(64, 5).to('client_a')
        
    def forward(self, first_hidden):
        last_hidden = self.hidden3(self.hidden2(self.hidden1(first_hidden))))
        return self.net2(last_hidden.to('client_a'))
        
# A and B initialize their model parameters
theta_a = client_a.init(32, 32)
theta_b = client_b.init(32, 32)

# clients load data
x_a = client_a.load_features('xa_location')
y = client_a.load_labels('y_location')
x_b = client_b.load_features('xb_location')
model = ToyModel()
loss = nn.CrossEntropyLoss()
optimizer = optim.SGD(model.parameters(), lr=0.001)
for iter in range(max_iter):

    # clients make private feature related forward computations using Python
    first_hidden = execute_algorithm_2(x_a, theta_a, x_b, theta_b)
    
    # server makes forward-backward computations in PyTorch
    optimizer.zero_grad()
    outputs = model(first_hidden)
    loss(outputs, y).backward()
    
    # clients make private feature related backward computations using Python
    first_hidden_gradient = first_hidden.grad.data
    update_client_models(first_hidden_gradient, x_a, theta_a, x_b, theta_b)
\end{lstlisting}
\caption{Example code of \modelname~for a neural network.}
\label{examcode}
\end{figure}
\section{Empirical Study}\label{exp}
In this section, we conduct comprehensive experiments to study the accuracy and efficiency of \modelname~by comparing with the state-of-the-art algorithm based privacy preserving DNN methods and cryptograph based privacy preserving DNN approaches. 

\subsection{Experimental Settings}

\nosection{Datasets} 
Note that \modelname~is a general privacy-preserving DNN model and can use applied into most scenarios where traditional DNN model works such as fraud detection, recommender systems, and link prediction tasks. 
To test the effectiveness of our proposed model, in our experiments, we choose two benchmark datasets, both of which are binary classification tasks. The first one is a fraud detection dataset \cite{dal2014learned}, where there are 28 features and 284,807 transactions. The other one is financial distress dataset \cite{findata}, where there are 83 features and 3,672 transactions. 
Since the financial distress dataset contains categorical features and DNN can only handle numerical input, we process these categorical features with one-hot encoding. 
After that, there are 556 features in total. 
We assume these features are hold by two parties, and each of them has equal partial features. Moreover, we randomly split the fraud detection dataset into two parts: 80\% as training dataset and the rest as test dataset. 
We also randomly split the financial distress dataset into 70\% and 30\%, as suggested by the dataset owner. 
We repeat experiments five times and report their average results. 

\nosection{Metrics} 
We adopt Area Under the receiver operating characteristic Curve (AUC) as the evaluation metric \cite{fawcett2006introduction}, since both datasets are binary classification tasks. 
In practice, AUC is equivalent to the probability that the classifier will rank a randomly chosen positive instance higher than a randomly chosen negative instance, and therefore, the higher the better. 

\nosection{Hyper-parameters} 
For the Fraud detection dataset, we use a multi-layer perception with 2 hidden layers whose dimensions are (8,8). We choose Sigmoid as the activation function \cite{jun1995sigmod} and use gradient descent as the optimizer. We set the learning rate to 0.001.
For the Financial distress dataset, we use a multi-layer perception with 3 hidden layers with dimensions (400, 16, 8), we choose Relu as the activation function \cite{HRrelu2000} in the last layer and Sigmoid function in the other layers, and set the learning rate to 0.006. 

\nosection{Comparison methods}
To study the effectiveness and efficiency of \modelname, we compare it with the following three kinds of approaches. 
\begin{itemize} [leftmargin=*] \setlength{\itemsep}{-\itemsep}
\item Plaintext Neural Network (\textbf{NN}) builds DNN using the plaintext data and therefore cannot protect data privacy. 
\item Split Neural Network (\textbf{SplitNN}) \cite{vepakomma2018split} builds privacy preserving DNN by split the computation graphs of DNN from algorithmic perspective, where each data holder trains a partial deep network model using its own features individually, and then the partial models are concatenated and sent to a server who has labels to train the rest of the mode. 
\item \textbf{SecureML} \cite{mohassel2017secureml} designs end-to-end privacy preserving DNN model using secret sharing protocols from cryptographic perspective. It also uses piece-wise functions or multi-nomials to approximate the non-linear active function in DNN. 
\end{itemize}

Moreover, our proposed \modelname~has two implementations, i.e., SS and HE, and therefore we have \textbf{\modelname-SS} and \textbf{\modelname-HE}. 

\subsection{Accuracy Comparison}\label{sec:exp:compare} 
We first study the accuracy (AUC) of \modelname. 
For accuracy comparison, we use \modelname~to denote both \modelname-SS and \modelname-HE, since they have the same AUC. Note that we use SGD as the optimizer during comparison. 

\subsubsection{Comparison results of two data holders}
We first assume there are only two parties, and report the comparison AUC performances on both datasets in Table \ref{compare_AUC}. 
From it, we can see that \modelname~achieves almost the same prediction performance as NN, and the differences come from the fixed-point representation of decimal numbers. 
We also observe that \modelname~has better performance than SplitNN and SecureML. This is because our proposed \modelname~uses cryptographic techniques for data holders to learn the first hidden layer collaboratively. In contrast, for SplitNN, the data holders learn partial hidden layer individually, which causes information loss. For SecureML, it has to approximate the multiply non-linear functions, which damages its accuracy. 

\begin{figure}[t]
\centering
\includegraphics[width=6cm]{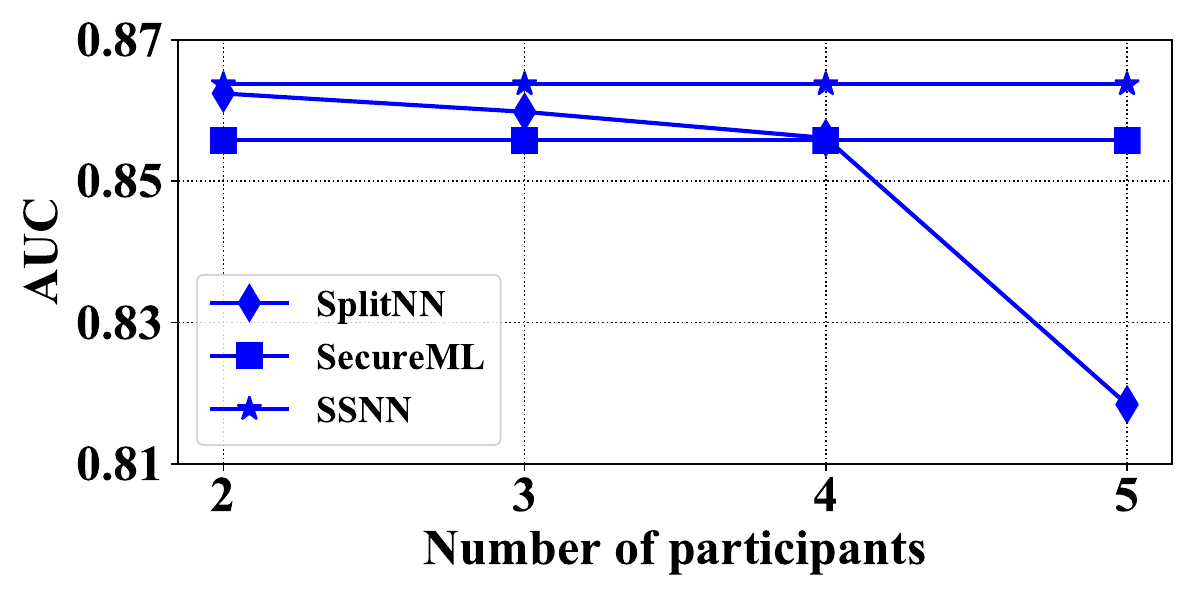}
\caption{Effect of the number of participants.}
\label{fig-effect-num}
\end{figure}

\subsubsection{Effects of the number of data holders}
We then study how the number of data holders affects each model's performance. Figure \ref{fig-effect-num} shows the comparison results with respect to different number of data holders, where we choose the fraud detection dataset. From it, we find that both SecureML and \modelname~achieve the same performance with the change of number of data holders. On the contrary, the performance of SplitNN tends to decline with the increase of number of participants. This is because, for both SecureML and \modelname, the data holders collaboratively learn all the layers or the first layer using cryptographic technique. As a contrast, for SplitNN, the data holders learn partial hidden layer individually, and the more data holders, the more information is lost. 

\subsubsection{Training and test losses}
\begin{figure}[t]
\centering
\subfigure [\emph{Training loss}]{ \includegraphics[width=4.2cm]{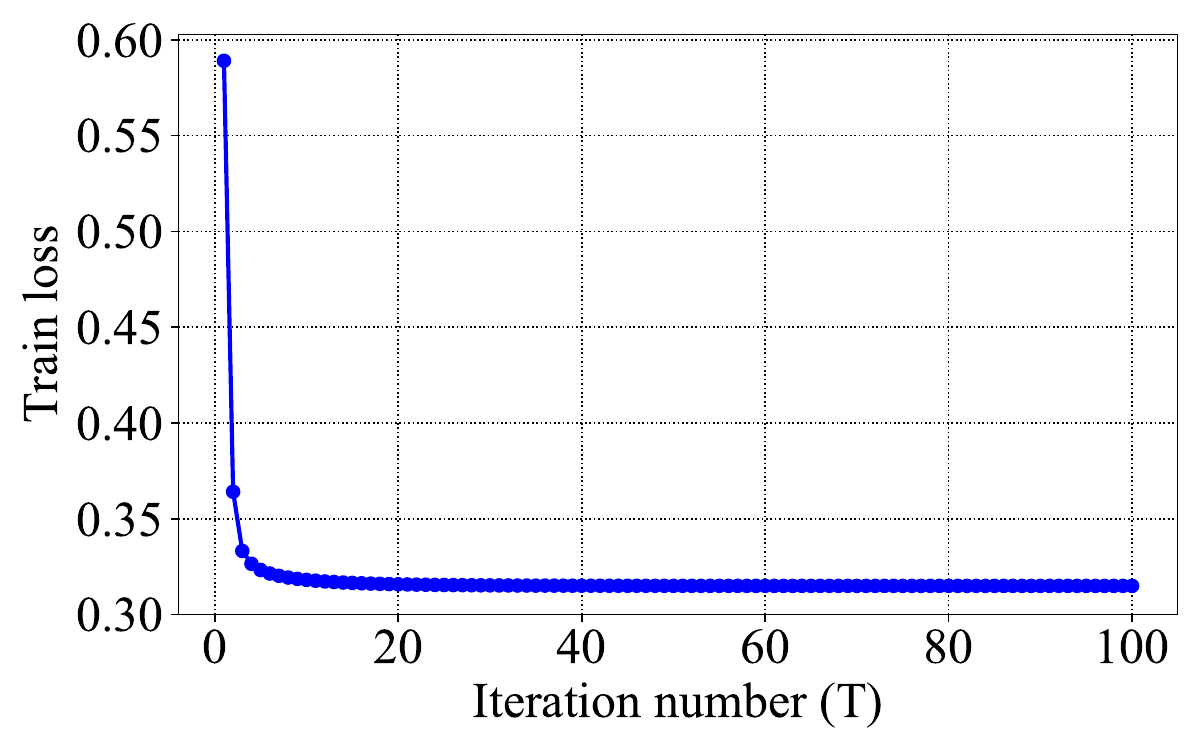}}~~~
\subfigure[\emph{Test loss}] { \includegraphics[width=4.2cm]{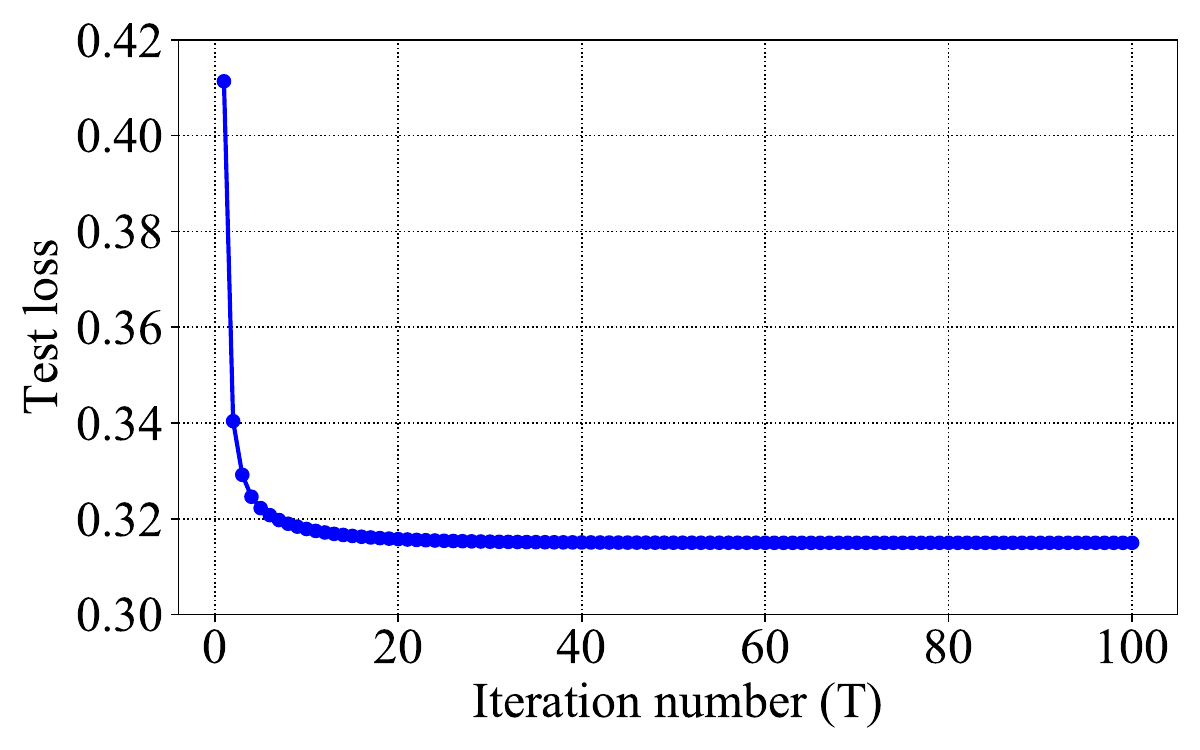}}
\caption{Average loss of \modelname~on fraud detection dataset.}
\label{loss1}
\end{figure}

\begin{figure}[t]
\centering
\subfigure [\emph{Training loss}]{ \includegraphics[width=4.2cm]{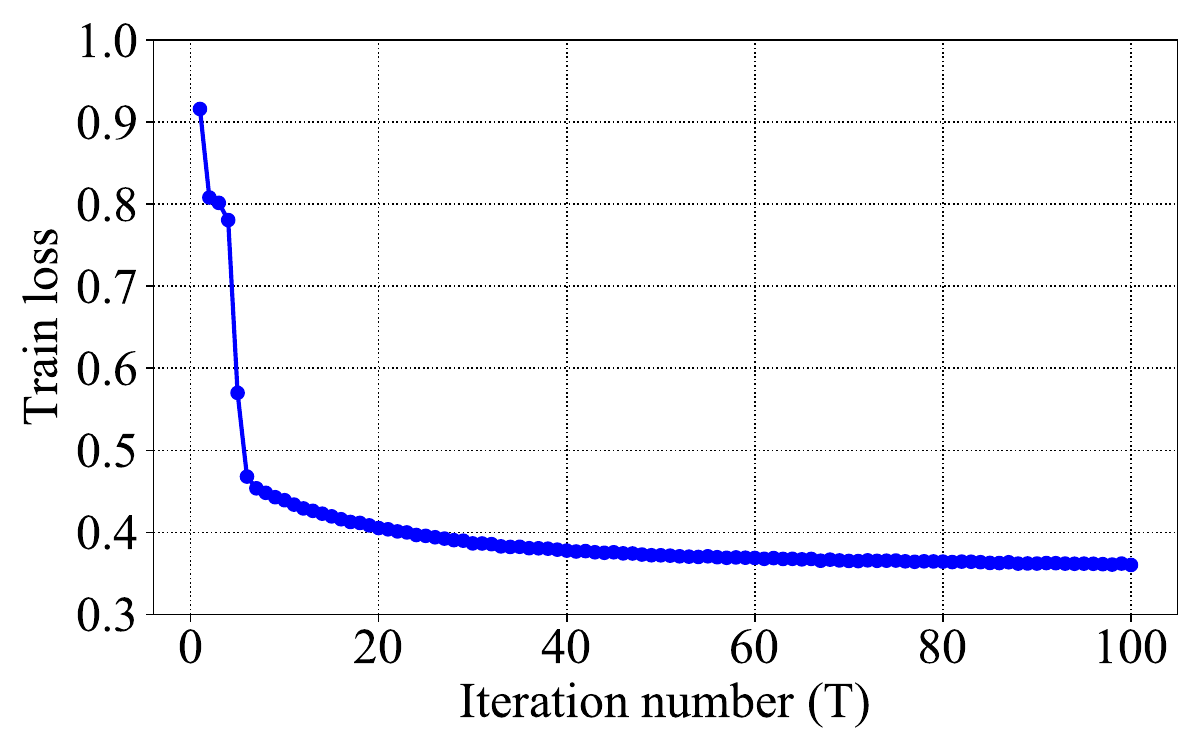}}~~~
\subfigure[\emph{Test loss}] { \includegraphics[width=4.2cm]{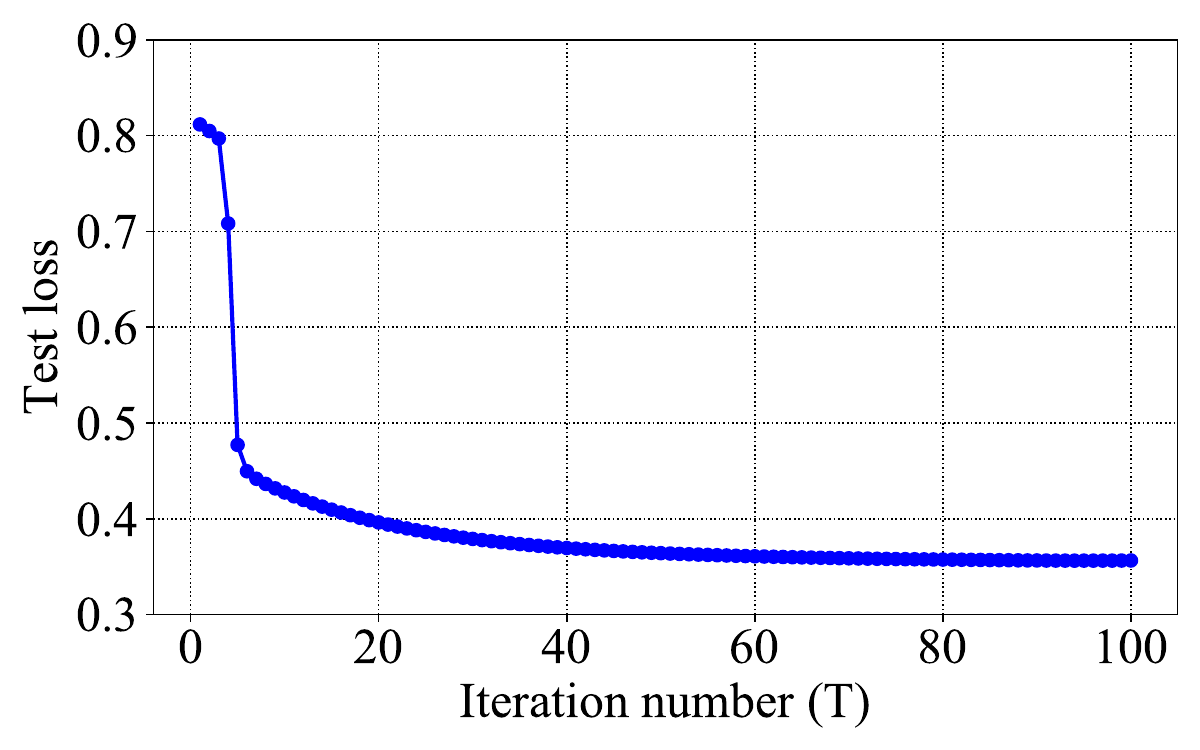}}
\caption{Average loss of \modelname~on financial distress dataset.}
\label{loss2}
\end{figure}

Besides, to study whether \modelname~has over-fitting problem, we study the average training loss and average test loss of \modelname~w.r.t. the iteration. 
We report the average training loss and average test loss of \modelname~on both datasets in Figure \ref{loss1} and Figure \ref{loss2}, respectively. 
From them, we can see that \modelname~converges steadily without over-fitting problem. 

The above experiments demonstrate that \modelname~consistantly achieves the best performance no matter what the nubmer of participants is, which shows its practicalness.

\begin{table}
\centering
\caption{Comparison results on two datasets in terms of AUC.}
\label{compare_AUC}
\begin{tabular}{|c|c|c|c|c|}
  \hline
  AUC & NN & SplitNN & SecureML & \modelname  \\
  \hline
  Fraud Detection & 0.8772 & 0.8624 & 0.8558 & 0.8637 \\
  \hline
  Financial Distress & 0.9379 & 0.9032 & 0.9092 & 0.9314 \\
  \hline
\end{tabular}
\end{table}

\subsection{Leakage Reduction of Hidden Features}
In this part, we empirically demonstrate the effectiveness of replacing SGD with SGLD to reduce the information leakage of hidden features (layers). To do this, we first introduce the property attack used in the prior work~\cite{ganju2018property}. This attack aims to infer whether a hidden feature has a specific property or not. For our specific task of fraud detection, we select `amount' as the target 
property. That is, we try to infer the amount of each transaction given the hidden features. 
For simplification, we change the value of `amount' to 0 or 1 based on its median, i.e., the values bigger than median are taken as 1, otherwise as 0. Therefore, the attack becomes a binary classification task. 

\nosection{Attack Model}
To quantify the information leakage of \modelname, we borrow the \textit{shadow training} attack technique from \cite{shokri2017membership}. First, we create a ``shadow model'' that
imitate the behavior of the \modelname, but for which we know the training datasets and thus the ground truth about `amount' in these datasets. We then train the attack model
on the labeled inputs and outputs of the shadow models. In our experiments, we assume the attacker somehow gets the `amount' (label) from the original dataset and the corresponding hidden features, with which the attacker tries to train the attack model. 
For this task,  we use the fraud detection dataset, from where we randomly split 50\% as the shadow dataset, 25\% as the training dataset and 25\% as the test dataset. 
Note that the data split of property attack is different from that of accuracy comparison in Section \ref{sec:exp:compare}. 
Here, we build a simple logistic regression model for property attack task. After this, to compare the effects of leakage reduction of different training methods, we perform the property attack on \modelname~trained by using SGD and SGLD and use the AUC to evaluate the attack performance. 

We report the comparison results in Table~\ref{tab:attack_results} using the fraud detection dataset. 
As Table~\ref{tab:attack_results} shows, SGLD can significantly reduce the information leakage compared with the conventional SGD in terms of AUC, i.e., 0.8223 vs. 0.5951. More surprisedly, SGLD also boosts the model performance in terms of AUC. Compared with SGD, SGLD has boosted the AUC value from 0.9118 to 0.9313. We infer that the performance boost can be caused by the regularization effect introduced by SGLD (i.e., SGLD can improve the generalization ability of the model).

\begin{table}[t]
\centering
\caption{Evaluation of information leakage on fraud detection dataset.}
\begin{tabular}{|c|c|c|}
\hline
Optimizer & Task AUC & Attack AUC \\
\hline
SGD       & 0.9118    & 0.8223      \\
\hline
SGLD      & 0.9313    & 0.5951 \\   
\hline
\end{tabular}
\label{tab:attack_results}
\end{table}

\subsection{Scalability Comparison}\label{sec-exp-speed}

We now study the scalability (efficiency) of our proposed \modelname, including its training time comparison with NN, SplitNN, and SecureML, the running time comparison of \modelname-SS and \modelname-HE, and the running time of \modelname~with different training data sizes, where the running time refers to the running time per epoch.

\subsubsection{Comparison of training time} 
First, we compare the training time of \modelname-SS with NN, SplitNN, and SecureML on both datasets.
The results are summarized in Table \ref{compare_time}, where we set batch size to 5,000 and the network bandwidth to 100Mbps.
From it, we find that NN and SplitNN are the most efficient ones since they do not involve any time-consuming cryptographic techniques. \modelname~is slower than NN and SplitNN since it adopts secret sharing technique for data holders to collaboratively calculate the first hidden layer. SecureML is much slower than \modelname~and is the slowest one since it uses secure multi-party computation techniques to calculate all the neural networks, and the speedup of \modelname~against SecureML will be more significant when the network structure is deeper. 
Note that both our proposed \modelname~and SecureML are secure against semi-honest adversaries. The results demonstrate the superior efficiency of \modelname. 

\begin{table}[t]
\centering
\caption{Comparison of training time per epoch (in seconds) on both datasets.}
\label{compare_time}
\begin{tabular}{|c|c|c|c|c|}
  \hline
  Training time & NN & SplitNN & SecureML & \modelname-SS  \\
  \hline
  Fraud detection & 0.2152 & 0.7427 & 960.30 & 37.22 \\
  \hline
  Financial distress & 0.0507 & 0.4541 & 751.29 & 21.84 \\
  \hline
\end{tabular}
\end{table}

\begin{figure}[t]
\centering
\subfigure [\emph{Fraud detection}]{ \includegraphics[width=4.2cm]{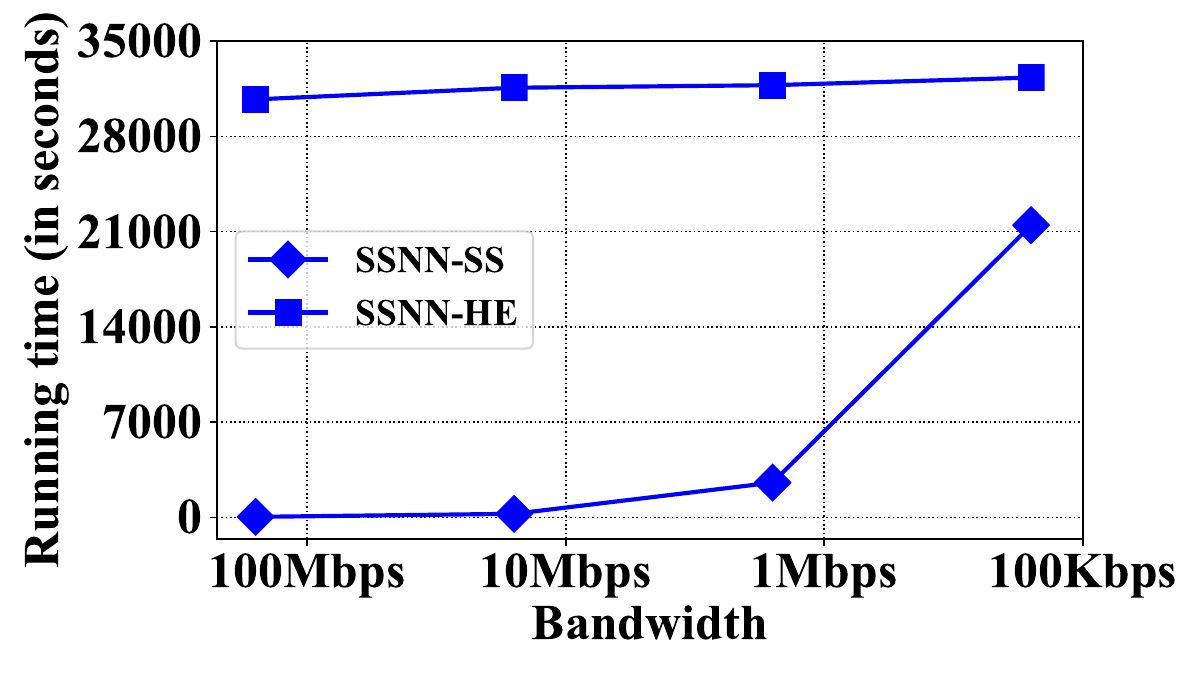}}~~~
\subfigure[\emph{Financial distress}] { \includegraphics[width=4.2cm]{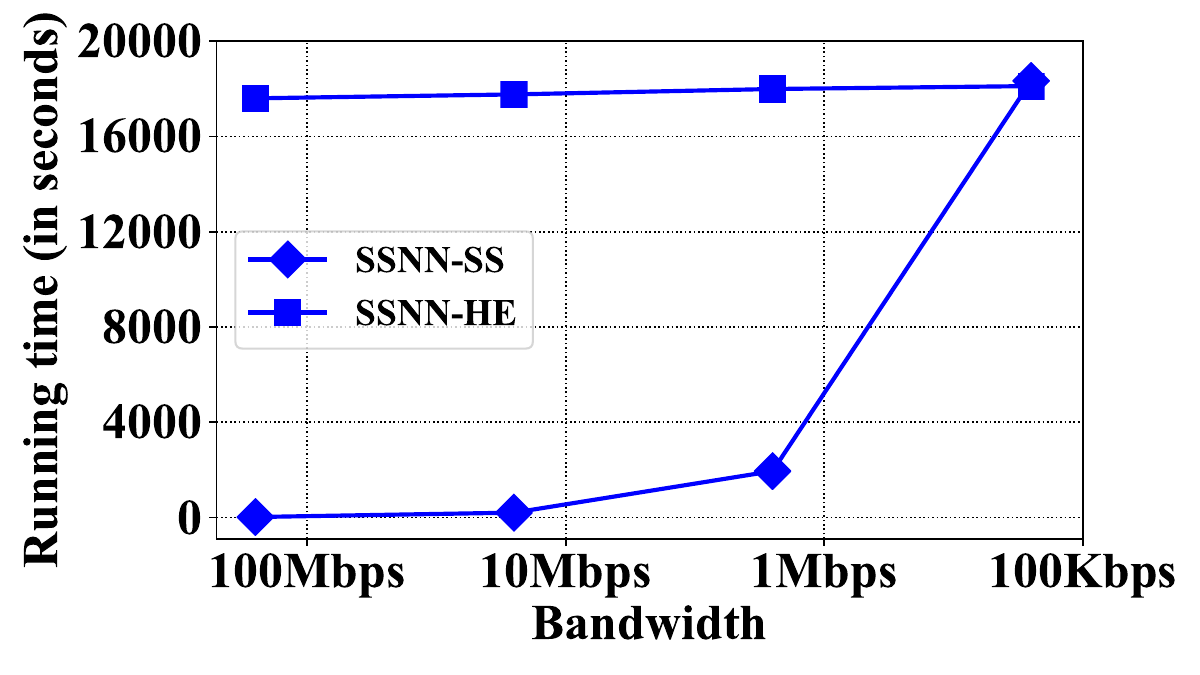}}
\caption{Efficiency comparison of \modelname-SS and \modelname-HE.}
\label{fig-ss-he-compare}
\end{figure}

\subsubsection{Comparison of \modelname-SS and \modelname-HE}
We then compare the efficiency of \modelname-SS and \modelname-HE, which implements \modelname~ using SS and HE, respectively. From Figure \ref{fig-ss-he-compare}, we find that the efficiency of \modelname-SS is significantly affected by network bandwidth, while \modelname-HE tends to be stable with respect to the change of bandwidth. When the network status is good (i.e., high bandwidth), \modelname-SS is much more efficient than \modelname-HE. However, \modelname-SS becomes less efficient than \modelname-HE when the network status is poor (i.e., low bandwidth), e.g., bandwidth=100Kbps on Financial distress dataset. The results indicate that our proposed two implementations are efficient and suitable for different network status.




\subsubsection{Running time with different batch size} 
Next, we study the running time of \modelname~with different batch size. 
We use the whole fraud detection dataset and report the running time of \modelname-SS in an epoch in Figure \ref{mpc_time_data} (a), where we use local area network. 
From it, we find that the running time of \modelname-SS decreases with the increase of batch size and tends to be stable gradually.  
This is because the number of interactions between participants will decrease with the increase of batch size. 

\subsubsection{Running time with different data size} 
\begin{figure}[t]
\centering
\subfigure [\emph{Effect of batch size on \modelname-SS}]{ \includegraphics[width=4.2cm]{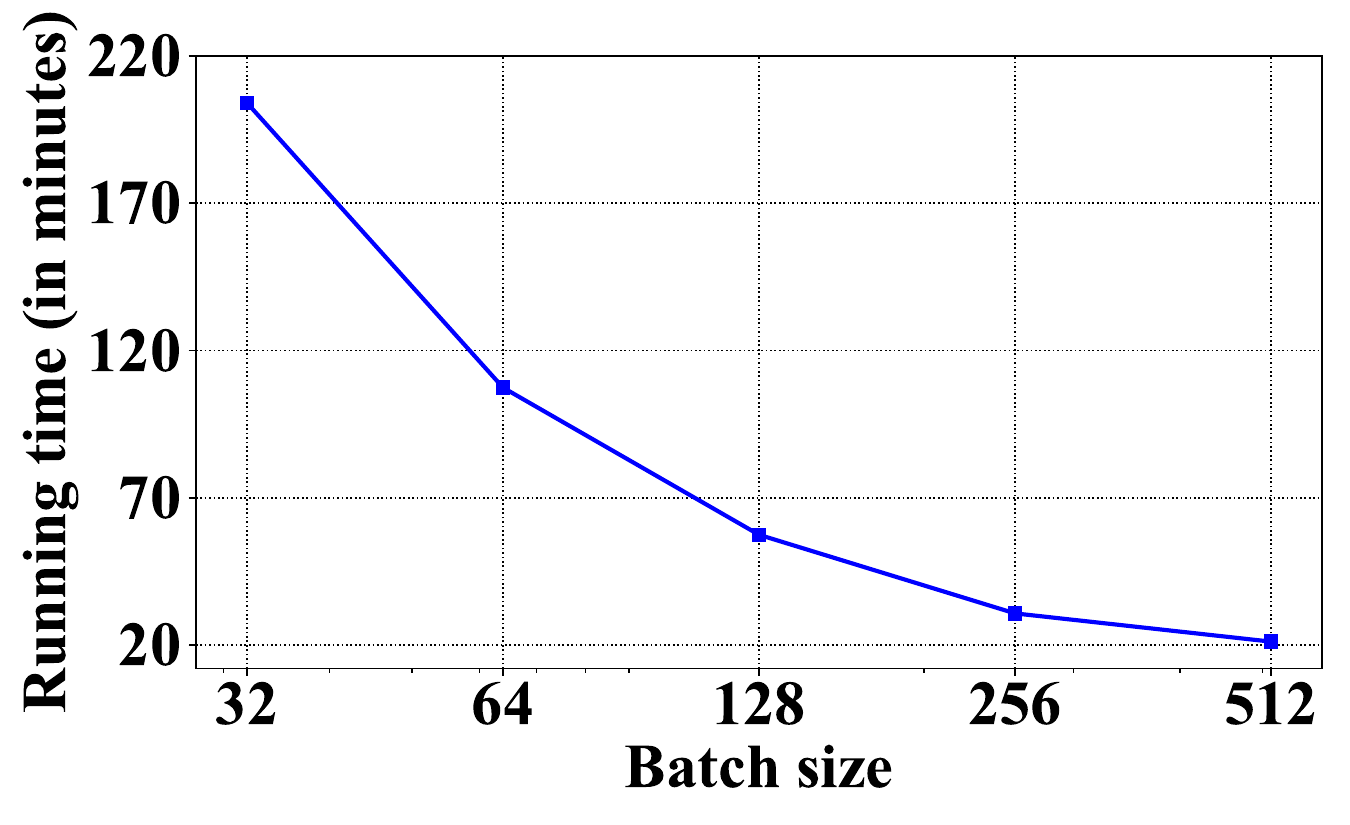}}~~~
\subfigure [\emph{Effect of data size on \modelname-SS}]{ \includegraphics[width=4.5cm]{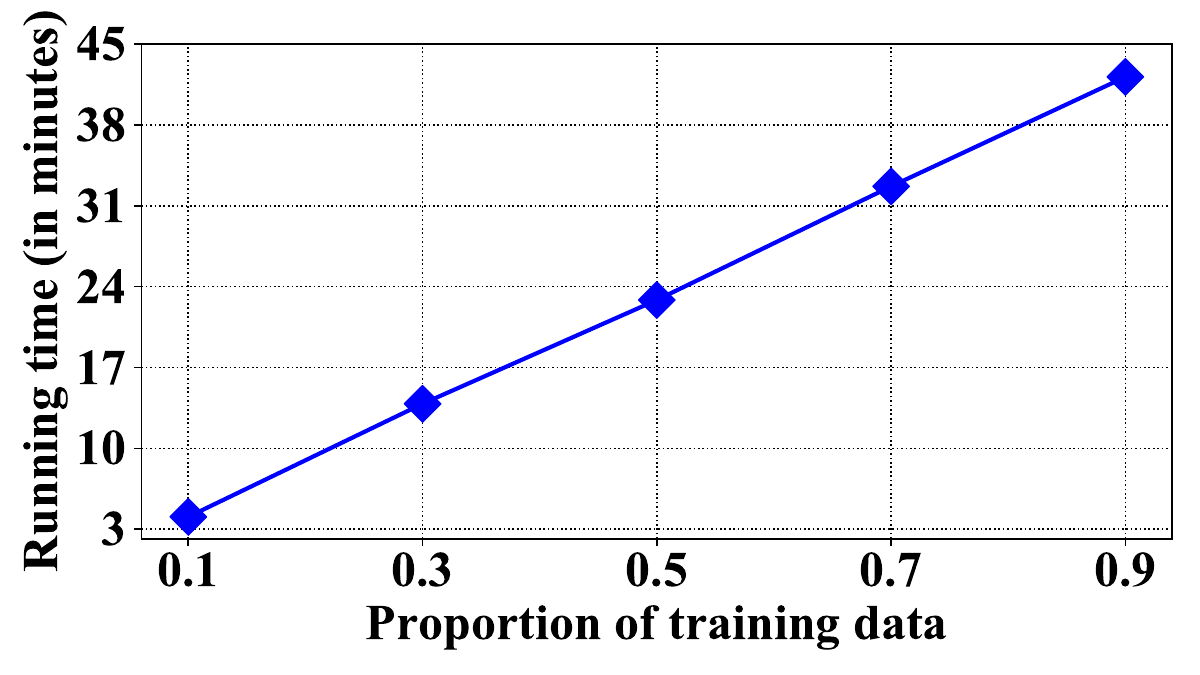}}~~~
\subfigure[\emph{Effect of data size on \modelname-HE}] { \includegraphics[width=4.5cm]{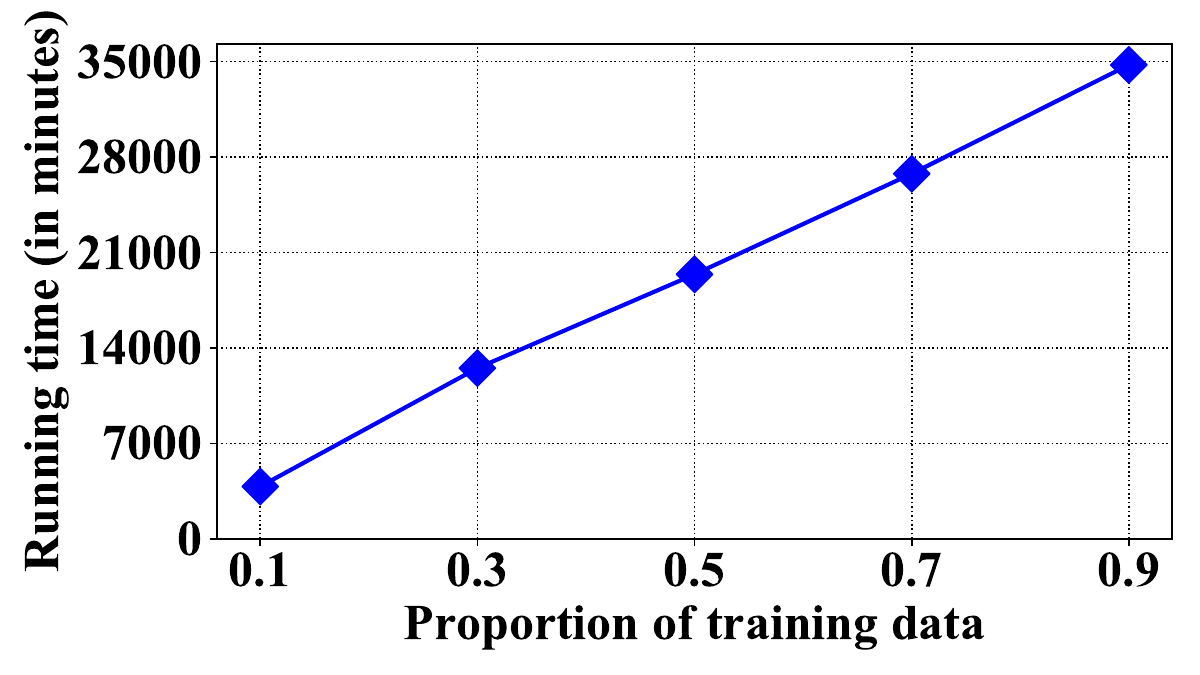}}
\caption{Effect of batch size and training data size on \modelname.}
\label{mpc_time_data}
\end{figure}

Finally, we study the running time of \modelname-SS and \modelname-HE with different data size, where fix the network bandwidth to 100Mbps.
We do this by varying the proportion of training data size using the fraud detection dataset, and report the running time of \modelname-SS and \modelname-HE in Figure \ref{mpc_time_data} (b) and (c). 
From them, we find that the running time of \modelname-SS and \modelname-HE scales linearly with the training data size. 
The results illustrate that our proposed \modelname~ can scale to large datasets.

\section{Conclusion and Future Work}\label{conlu}
In this paper, we have proposed \modelname~---~a scalable privacy preserving deep neural newtwork learning. 
Our motivation is to design \modelname~from both algorithmic perspective and cryptographic perspective. 
From algorithmic perspective, we split the computation graph of DNN models into two parts, i.e., the private data related computations that are performed by data holders, and the rest heavy computations which are delegated to a semi-honest server with high computation ability. 
From cryptographic perspective, we proposed two kinds of cryptographic techniques, i.e., secret sharing and homomorphic encryption, for the isolated data holders to conduct private data related computations privately and cooperatively. 
We implemented \modelname~in a decentralized setting and presented its user-friendly APIs. 
Our model has achieved promising results on real-world fraud detection dataset and financial distress dataset. 
In the future, we would like to deploy our proposal in real-world applications.

\begin{acks}
This work was supported in part by the National Key R\&D Program of China (No.2018YFB1403001).
\end{acks}

\bibliographystyle{ACM-Reference-Format}
\bibliography{cement}

\end{document}